\documentclass{article}
\usepackage{ijcai24}

\usepackage{times}
\usepackage{multirow}
\usepackage{soul}
\usepackage{url}
\usepackage[hidelinks]{hyperref}
\usepackage[utf8]{inputenc}
\usepackage[small]{caption}
\usepackage{graphicx}
\usepackage{amsmath}
\usepackage{amsthm}
\usepackage{booktabs}
\usepackage{algorithm}
\usepackage{algorithmic}
\usepackage[switch]{lineno}
\usepackage[dvipsnames]{xcolor}
\usepackage{natbib}
\usepackage{enumitem, tabularx}
\usepackage{amsfonts}

\usepackage{latexsym}

\title{Visual Attention Prompted Prediction and Learning}

\author{
Yifei Zhang$^1$\footnote{Code and tools available at \url{https://github.com/yifeizhangcs/visual-attention-prompt}} \and
Bo Pan$^1$ \and
Siyi Gu$^2$ \and
Guangji Bai$^1$ \and 
Meikang Qiu$^3$ \and \\
Xiaofeng Yang$^1$ \and
Liang Zhao$^1$\\
\affiliations
$^1$Emory University\\
$^2$Stanford University\\
$^3$Augusta University\\
\emails
\{yifei.zhang2, bo.pan, guangji.bai, xyang43, liang.zhao\}@emory.edu,
sgu33@stanford.edu,
qiumeikang@yahoo.com
}

\begin{document}

\maketitle

\begin{abstract}
Visual explanation (attention)-guided learning uses not only labels but also explanations to guide model reasoning process. While visual attention-guided learning has shown promising results, it requires a large number of explanation annotations that are time-consuming to prepare. However, in many real-world situations, it is usually desired to prompt the model with visual attention without model retraining. For example, when doing AI-assisted cancer classification on a medical image, users (e.g., clinicians) can provide the AI model with visual attention prompt on which areas are indispensable and which are precluded. Despite its promising objectives, achieving visual attention-prompted prediction presents several major challenges: 1) How can the visual prompt be effectively integrated into the model's reasoning process? 2) How should the model handle samples that lack visual prompts? 3) What is the impact on the model's performance when a visual prompt is imperfect? This paper introduces a novel framework for attention-prompted prediction and learning, utilizing visual prompts to steer the model's reasoning process. To improve performance in non-prompted situations and align it with prompted scenarios, we propose a co-training approach for both non-prompted and prompted models, ensuring they share similar parameters and activations. Additionally, for instances where the visual prompt does not encompass the entire input image, we have developed innovative attention prompt refinement methods. These methods interpolate the incomplete prompts while maintaining alignment with the model's explanations. Extensive experiments on four datasets demonstrate the effectiveness of our proposed framework in enhancing predictions for samples both with and without prompt.
\end{abstract}

\section{Introduction}
\label{sec:intro}
The ``black box'' nature of deep learning models often obscures the decision-making process in AI, leading to the emergence of Explainable AI (XAI) aimed at demystifying the rationale behind models~\citep{adadi2018peeking}. Techniques like CAM, Grad-CAM, and integrated gradients, pivotal in XAI, produce saliency maps highlighting the model's focus areas in the input data~\citep{zhou2015learning, selvaraju2017grad,qi2019visualizing,bai2023saliency}. While XAI has advanced in explaining model reasoning, the ultimate aim extends beyond this. It is crucial to leverage XAI to enhance the reasoning and predictive capabilities of Deep Neural Networks (DNNs). Key challenges include incorporating human insight into the model's reasoning and ensuring that such insights positively impact future predictions.

\begin{figure}[h]
    \centering
    \includegraphics[width=\linewidth]{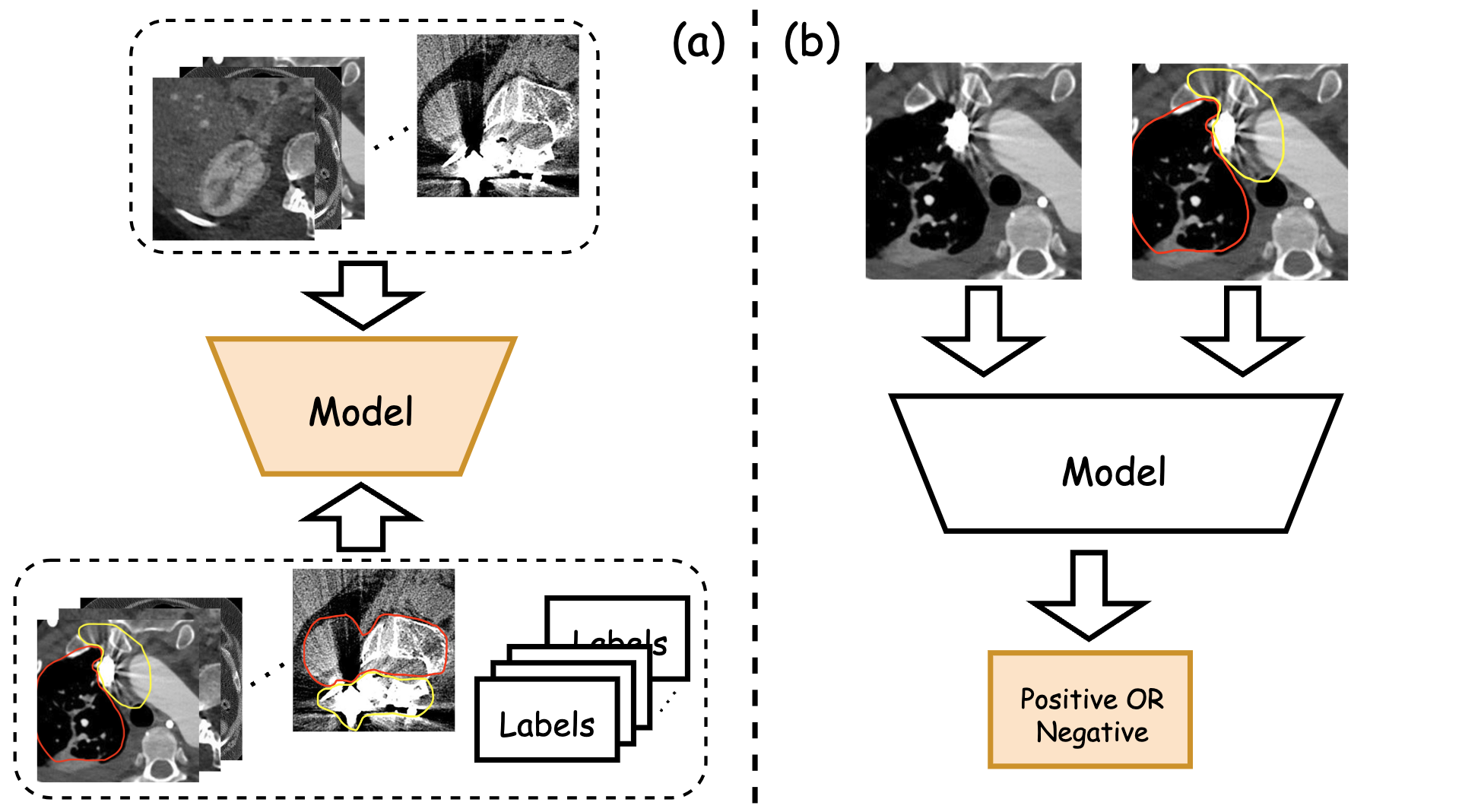}
    \caption{The comparison between attention-guided learning and attention-prompted prediction. (a) explanation-guided learning requires many user-annotated explanations to train the models, while (b) attention-prompted prediction enables users to directly guide the model's prediction process by telling the model which areas are ``indispensable'' (areas in red that look suspicious), ``precluded'' (areas in yellow that contain artifacts), and ``undecided'' (other areas).}
    \label{fig:example}
\end{figure}

Explanation-guided learning in natural language processing (NLP) task has been extensively researched, with studies such as \cite{hsieh2023distilling}, \cite{raffel2020exploring}, \cite{narang2020wt5} extracting rationales as additional supervision, alongside label supervision, for training language models. In contrast, the visual domain remains not well explored. In recent years, there has been emerging research in visual explanation-guided learning. Studies like \cite{chen2020air}, \cite{zhang2023magi}, \cite{shen2021human}, and \cite{gu2023essa} have begun utilizing human-annotated attention maps to guide the reasoning process of DNNs in Computer Vision (CV) tasks by simultaneously minimizing prediction errors and the disparity between the model's reasoning and the human-crafted true reasoning within the training set, as shown in Figure~\ref{fig:example}(a). However, visual explanation-guided learning is labor-intensive, time-intensive, and computationally expensive due to the need for a large amount of human-annotated attention maps in training data. In many practical applications, users may have easy and quick (high-level) guidance toward the model for a prediction, e.g., which areas are roughly more important are which are not. For instance, in cancer imaging, clinicians can quickly sketch out areas that are interesting or irrelevant in terms of determining whether the image indicates ``cancerous'', as shown in Figure~\ref{fig:example}(b). Hence, there's a need for an efficient way for them to incorporate these cues to assist the model's decision-making process. Despite many existing works about textual prompts as~\cite{oymak2023role}, \cite{liu2023kept}, and \cite{ye2022ontology}, how to prompt in visual space is much less underexplored and the focus of this paper. 

Despite the interestingness and significance, achieving visual attention-prompted prediction requires working on visual patterns on images that trigger unique unsolved challenges~\textbf{1) How to incorporate the visual attention prompt into the model's prediction?} Traditional classifiers only take images for decision-making, but how to embed the prompt such that it can guide the model reasoning as instructed in the visual attention prompt is crucial and very challenging.~\textbf{2) What if some samples do not have visual attention prompts?} Many samples may not come with prompts, but can their predictions still get some guidance from the samples with prompts? And how? and~\textbf{3) How to handle the incomplete visual attention prompts?} In practice, it is much easier and quicker to provide an imperfect prompt which may not cover the whole image but just its regions where the user can quickly have some intuition. How can we address the incompleteness and sufficiently utilize it?

To tackle the aforementioned challenges, we introduce the Visual Attention-Prompted Prediction and Learning framework. To address the first challenge, we proposed an attention-prompted prediction framework for integrating visual attention prompts into the model's decision-making process. To counter the second challenge, we developed an attention-prompted co-training mechanism that distills knowledge from the prompted model to the non-prompted model, thereby enhancing future prediction performance for samples without provided prompts. Finally, to tackle the third challenge, we proposed a novel architecture to achieve attention prompt refinement by automatically learning the saliency of ``undecided'' areas for each pixel. In summary, the main contributions of this paper are as follows: 1) A new framework designed to integrate visual attention prompts into the model's decision-making process; 2) A new attention-prompted co-training algorithm developed to improve predictions for samples without attention prompts by distilling knowledge from the attention-prompted model to the non-attention-prompted model; 3) A novel architecture to refine the incomplete visual attention prompts by automatically learning the saliency of ``undecided'' areas at the pixel level; 4) Comprehensive experiments on four datasets to demonstrate the effectiveness of our framework in enhancing model predictability for image classification tasks.


\section{Related Work}
\noindent\textbf{Privileged information} The learning with privileged information paradigm, as proposed by Vapnik~\citep{vapnik2015learning}, introduces a ``teacher'' that provides additional information to a ``student'' model during the learning process. The underlying idea is that the teacher’s extra explanations help the student develop a more effective model. Lopez et al.~\citep{lopez2015unifying} integrated the concepts of distillation and privileged information into 'generalized distillation,' a framework for learning from multiple machines and data representations. Garcia et al.~\citep{garcia2018modality} introduced a method for multimodal video action recognition. Pan et al.~\citep{pan2024distilling} applied privileged information from Large Language Models (LLMs) to Graph Learning.

\noindent\textbf{Attention-guided Learning} The integration of human knowledge into interpretable models has been extensively studied in CV and NLP tasks. Attention maps, such as those generated by Grad-CAM~\citep{montavon2019layer, selvaraju2017grad} or intrinsic attention mechanism~\citep{vaswani2017attention}, have been utilized as supervision signals. These signals are aligned with prediction loss to further enhance model performance~\citep{shen2021human, camburu2018snli, hajialigol2023xai}. HAICS~\citep{shen2021human} proposed a conceptual framework for image classification that incorporates human annotations in the form of scribble annotations as the attention signal. RES~\citep{gao2022res} developed a novel objective to handle inaccurate, incomplete, or inconsistently distributed issues of explanation.

\noindent\textbf{Attention Prompt} Prompting, originating from NLP tasks as shown by Devlin et al.~\citep{devlin2018bert}, has been adapted for CV applications~\cite{dosovitskiy2020image}. Oymak et al.~\citep{oymak2023role} examine prompt-tuning for one-layer attention architectures in contextual mixture models. Jia et al.~\citep{jia2022visual} introduced Visual Prompt Tuning (VPT) to add adaptable prompt tokens to the Vision Transformer (ViT) model's patch tokens. Paiss et al.~\citep{paiss2022no} proposed an explainability-based method for improving one-shot classification rates. Finally, Li et al.~\citep{li2023boosting} developed ``saliency prompts'' from saliency masks to highlight potential objects in scenes.

\begin{figure*}
    \centering
    \includegraphics[width=0.8\linewidth]{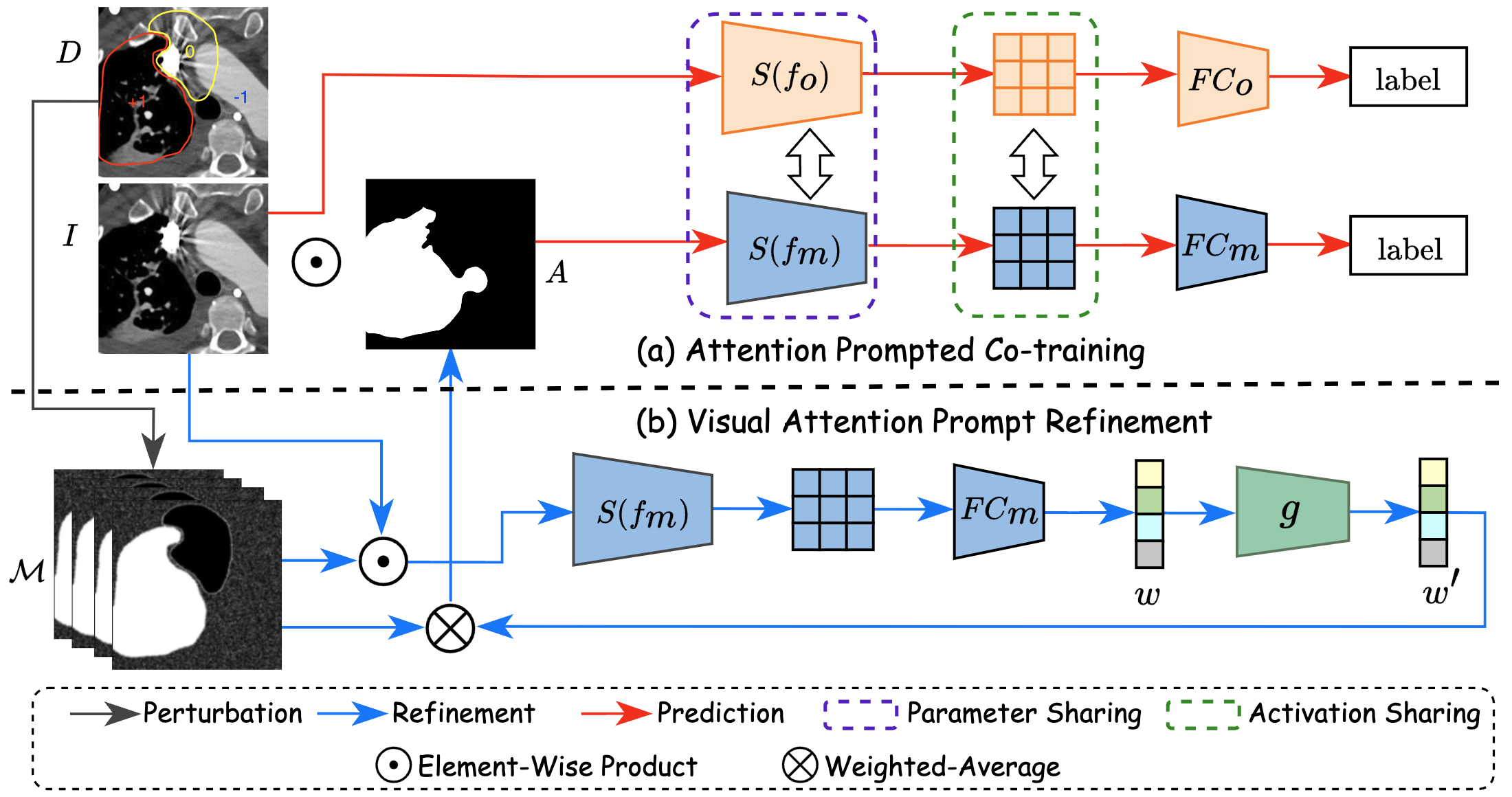}
    \caption{Illustration of the Visual Attention Prompted Prediction and Learning Framework: (a) depicts our proposed Attention-Prompted Co-Training Mechanism, while (b) outlines the proposed Visual Attention Prompt Refinement Architecture.}
    \label{fig:framework}
\end{figure*}

\section{Problem Formulation}

In the context of visual attention-prompted learning and prediction, we consider the samples from a dataset $\mathcal{U}$ to be provided in pair as $(I, D, y) \in \mathcal{U}$, where $I \in \mathbb{R}^{C \times H \times W}$ represents the original image, with $C$, $H$, and $W$ denotes the number of channels, height, and width, respectively, $y$ is the class label, and visual attention prompt for the image corresponding to its class label is denoted by $D \in \mathbb{R}^{H \times W}$, with dimensions identical to the original image but with one channel. Attention prompt $D$ serves as an instruction or signal indicating which parts of the input image are particularly relevant or should be prioritized. The visual attention-prompted prediction process of model $f$ can be denoted as $f: (I, D) \mapsto y$. 

Despite the merit of the potential of embedding the visual attention prompt into a prediction to guide it better, achieving it requires tackling nontrivial technical challenges in three key aspects. Firstly, how to use the prompt to guide the model's reasoning so that it can attend to the right places? Secondly, image samples $I$ may not always come with prompts $D$, but can they benefit from the prompts of the others? Moreover, it is usually much more efficient to provide incomplete prompts covering only part of the image, but how to handle such incompleteness?

\section{Methodology}
To address the challenges outlined above, we propose a new framework for visual attention-prompted prediction and learning. Section~\ref{sec:framework} introduces the overall framework we propose. Section~\ref{sec:refinement} details the attention prompts refinement architecture. Finally, in Section~\ref{sec:training}, we propose the attention-prompted co-training mechanism.

\subsection{Overall framework}
\label{sec:framework}
In our Visual Attention Prompted Prediction Framework, attention prompts are employed to effectively guide the reasoning process of the AI model. This is achieved by utilizing the attention prompt to mask the original image. Consequently, the model's decision-making process is influenced by areas categorized as ``indispensable,'' ``precluded,'' and ``undecided,'' ensuring that these guided insights are integral to the AI's analytical procedures. In the left of Figure~\ref{fig:framework}(a), ``indispensable'' and ``precluded'' areas are shown in red and yellow, respectively, with the remaining areas being ``undecided'' ones. To address the incompleteness of the prompt, we will respect certain parts that are either ``indispensable'' or ``precluded'', but impute the undecided part by eliciting the model's explanation of the reasoning process and aligning it with the given attention prompt. As illustrated in Figure~\ref{fig:framework}(b), the given incomplete prompt is initially inputted into the Visual Attention Prompt Refiner. This step serves to activate and constrain the generation of post-hoc explanations. Subsequently, these elicited post-hoc explanations are refined and utilized to mask the input image, thereby facilitating the attention-prompted prediction process. The details are elaborated on in Section~\ref{sec:refinement}. As illustrated in Figure~\ref{fig:framework}(a), to handle samples without prompts and allow them to benefit from those with prompts, we propose another non-prompted model, \( f_o \), that takes only the image. This model is co-trained with the prompted model \( f_m \) by aligning both the model parameters and activations, as indicated by the dashed rectangles in the figure. \( S(f_o) \) and \( S(f_m) \) denote the sets of model architectures before the fully-connected layers, while \( FC_o \) and \( FC_m \) represent the fully-connected layers of two models respectively. The details will be elaborated in Section~\ref{sec:training}.


\subsection{Visual Attention Prompt Refinement}
\label{sec:refinement}
In this section, we propose an attention-prone refinement architecture. First, we introduce prompt-guided incomplete prompt refinement architecture. Second, we present an adaptively learnable mask aggregation method.

\subsubsection{Prompt-Guided Incomplete Prompt Refinement}
To refine the incomplete prompt that contains some areas ``undecided'' to be important to the prediction, we propose a prompt-guided incomplete prompt refinement architecture. This approach aims to learn the saliency of undecided areas by aligning the given incomplete prompt with post-hoc explanations. Consider a prompt \( D \in \mathbb{R}^{H \times W} \) that includes annotations distinctly highlighting indispensable, precluded, and undecided areas, we aim to leverage the post-hoc explanations of the prediction for the image prompted with this incomplete prompt, to refine the prompt. However, those gradient-based explainers are usually time-consuming due to the involvement of back-propagation during explanation generation. Therefore, post-hoc explanations obtained without backpropagation such as perturbation-based methods that can directly predict the importance of perturbations to compose explanations are preferred here. 

To be specific, each element \( \lambda \in D \) corresponds to a pixel in the image and takes a value from the set \( \{-1, 0, +1\} \). Here, \( \lambda=0 \) indicates that the pixel is precluded, \( \lambda=+1 \) signifies that the pixel is indispensible, and \( \lambda=-1 \) denotes that the pixel is undecided. To generate the post-hoc explanation guided by prompt $D$, we first randomly perturb $D$ as $P(D)$ to generate $N$ binary masks $\mathcal{M}=\{P^{(i)}(D)=M_i\}_{i=1}^{N}$, where $P^{(i)}(\cdot)$ denotes the $i$-th preturbation process, by setting each pixel with a value equal to $-1$ to $+1$ with probability $p$ and to $0$ otherwise. After obtaining the perturbed masks, we first perform an element-wise multiplication of the original image with each randomly perturbed mask to obtain the masked image. Subsequently, we compute the confidence scores for each mask in \( \mathcal{M} \) as \( \{(I \odot M_i)\}_{i=1}^{N} \), where \( I \) is the image and \( \odot \) represents the element-wise multiplication operation. We then calculate the confidence score for each masked image relative to its corresponding class label by \( w_i = \operatorname{softmax}_{k} f_m(I \odot M_i), i = 1, \ldots, N \), where \( k \) represents the index of the label \( y \) and \( \operatorname{softmax}_{k} \) denotes the output of the softmax layer corresponding to the \( k \)-th class. Upon calculating the confidence scores for each perturbed mask, we applied weighted averaging and normalized it with the expected pixel value \( N \cdot p \) for aggregation. The process resulted in the refined prompt, denoted as \( A \). To summarize, the above computation process can be represented as
\begin{equation}
\label{eq:old_aggregation}
A = \frac{1}{N \cdot p} \sum_{i=1}^{N} f_m(I \odot P^{(i)}(D)) \cdot P^{(i)}(D)
\end{equation}
where the refined prompt is obtained by learning the saliency of ``undecided'' areas to align with the post-hoc explanation, which is guided by the incomplete prompt.
 
\subsubsection{Adaptively Learnable Masks Aggregation}
\label{sec:aggregation}
Aggregating masks by the confidence score can ensure the relationship between confidence and importance, as a higher confidence score indicates greater importance in the aggregation process. However, the confidence score may not necessarily accurately quantify the importance. To overcome this, we require a function that can adaptively learn to quantify the importance score accurately. In the following, we propose to learn such a weight-learning function by satisfying the following inherent constraints of the confidence score:
\begin{itemize}[leftmargin=*, topsep=0 pt, partopsep=0 pt, itemsep=0 pt]
    \item \textit{(Monotonicity)}: The function needs to be monotonically non-decreasing, ensuring masks with higher confidence scores receive greater importance in the aggregation.
    \item \textit{(Endpoint-preserving)}: The function must ensure that inputs of 0 and 1 yield outputs of 0 and 1, respectively. This respects the input range's limits, preserving the key relationship between confidence scores and weights, and mitigates the effects of extremely large or negative values.
\end{itemize}

\begin{figure}[h]
    \centering
    \includegraphics[width=0.85\linewidth]{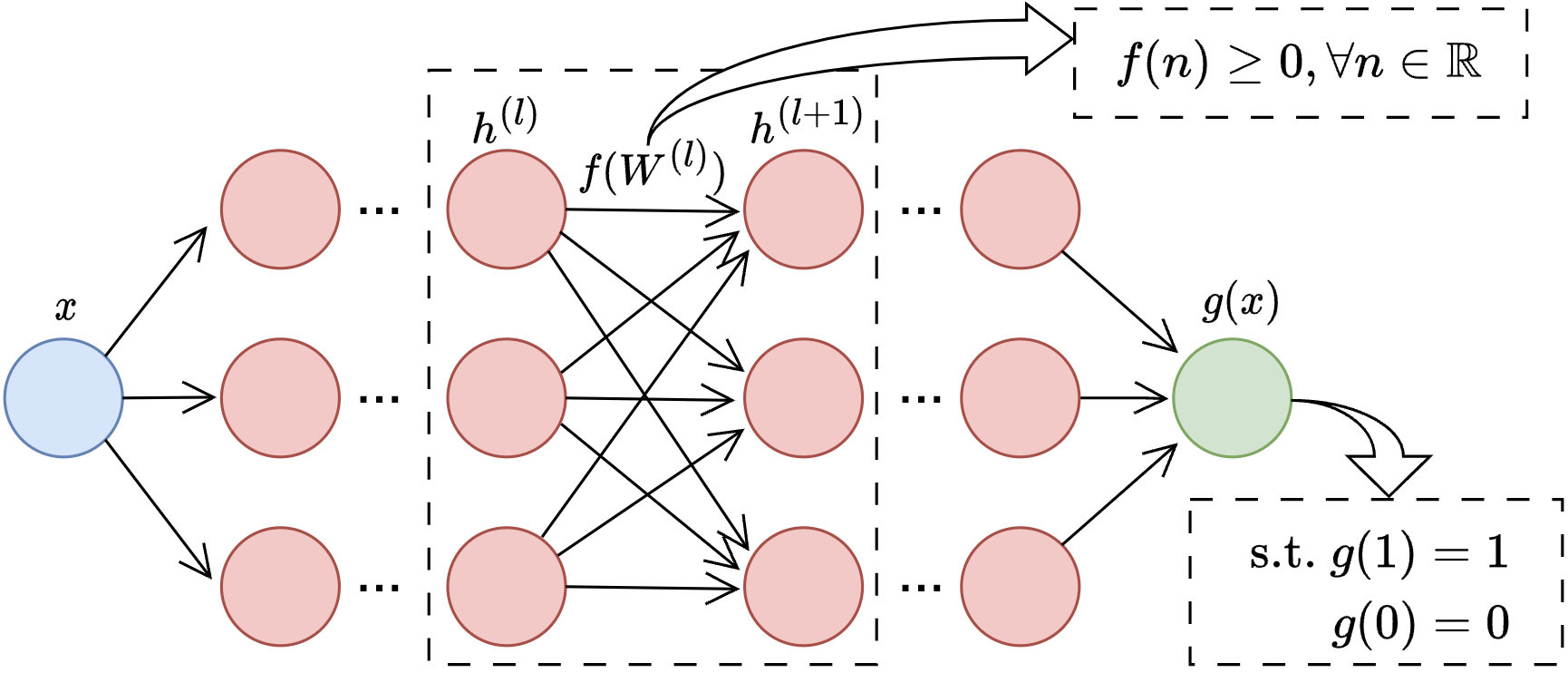}
    \caption{Visualization of proposed weights-learning function based on constrained MLP architecture.}
    \label{fig:mlp}
\end{figure}

To achieve the formulation and learning of such as weight-learning function, we build a constrained multilayer perceptron (MLP) \( g(\cdot) \), to quantify the importance scores. To be more specific, as shown in Figure~\ref{fig:mlp}, ensuring the \textit{monotonically non-decreasing} behavior, \( g(\cdot) \) is achieved by the principle that the weight matrix must consist of non-negative entries. This crucial constraint is enforced by applying weights derived from a function with a range of positive numbers~\citep{nguyen2023mononet}. To satisfy the \textit{endpoint-preserving} property, two specific measures have been taken: 1) To guarantee that the function produces an output of 0 when the input is 0, the bias term has been eliminated from all layers of \( g(\cdot) \). Additionally, an activation function $\sigma(\cdot)$ has been selected such that $\sigma(0)=0$, and 2) To ensure that the function yields an output of 1 when the input is 1, a constraint term $g(1) = 1$ is incorporated into the training phase. In details, the mapping between layer $k$ and layer $k+1$ of \( g(\cdot) \), which is depicted in Figure~\ref{fig:mlp}, is as
\begin{equation}
\begin{aligned}
h^{(k+1)}=\sigma(\phi (W^{(k)}) h^{(k)}), k=1,\ldots,L
\end{aligned}
\end{equation}
where \(\phi\) is a function that has a range within the positive numbers and can be exemplified by functions like the exponential function or a translated hyperbolic tangent, and \(L+1\) denotes the total number of layers in the model. With proposed function \( g(\cdot) \), the mask aggregation function, as described in Equation~\ref{eq:old_aggregation}, can be redefined as
\begin{equation}
\begin{aligned}
A = \frac{1}{N\cdot p} \sum_{i=1}^N g(f_m(I\odot P^{(i)}(D))) \cdot P^{(i)}(D)
\end{aligned}
\end{equation}
where the refined prompt is obtained by aggregating masks with the learned weights, as shown in Figure~\ref{fig:framework} (b).

\subsection{Attention Prompted Co-training}
\label{sec:training}
In this section, we delineate the attention-prompted co-training mechanism. Initially, the framework for parameter-sharing and co-activation is introduced. This is succeeded by a detailed exposition of the learning algorithm, which is predicated on an alternating optimization training strategy.

\subsubsection{Parameter-Sharing and Co-Activation Framework}

In instances where a prompt is unavailable, the proposed function \( f_m \), which necessitates a prompt, becomes inapplicable. Under such circumstances, an alternative predictor, designated as \( f_o \), is employed. This predictor operates exclusively based on the image data. Rather than learning \( f_m \) and \( f_o \) independently, we investigate their interrelation. For each training sample accompanied by a prompt, both \( f_m \) and \( f_o \) are executed, with the anticipation that \( f_o \) will mirror the correct reasoning dictated by \( f_m \) and guided by the prompt. This approach not only aims to encourage two models to have similar parameters but also to foster similar activation patterns. For instance, if a prompt directs \( f_m \) to disregard the region of an artifact in the image, which results in no activations in that region,  \( f_o \) should also follow it by disregarding this region. This alignment of activation patterns enables the transfer of knowledge from \(f_m \) to \( f_o \) in a joint parameter-sharing and co-activation framework. To concretize this concept, we propose the integration of two regularization terms: model parameter-sharing regularization, and co-activation regularization, which collectively embody this approach as
\begin{equation}
\mathcal{L_{\text{Param}}}(\theta_{f_m}, \theta_{f_o})=\|W_{f_o}-W_{f_m}\|_F^2
\label{eq:param}
\end{equation}
\begin{equation}
\mathcal{L_{\text{Activ}}}(\theta_{f_m}, \theta_{f_o})=\|S(f_o(I))-S(f_m(I \odot A))\|_F^2
\label{eq:feature}
\end{equation}
where $W_{f_m}$ and $W_{f_o}$ represent the convolutional layer parameters of two models and \(\|\cdot\|_F^2\) represents the squared Frobenius norm~\citep{279274}. Furthermore, the cross-entropy loss associated with the predictions made by the two models, when compared to the target, can be articulated as
\begin{equation}
\begin{aligned}
\mathcal{L}_{\text{Pred}} = -\sum_{i=1}^K \sum_{a=1}^C \hat{y}_{ia} \left(\log \left(p_{m}^{ia}\right) + \log \left(p_{o}^{ia}\right)\right)
\label{eq:pred}
\end{aligned}
\end{equation}
where $\hat{y}_{ia}$ is the ground truth label for class $a$ of the $i^{th}$ data point in one-hot encoding, $p_{m}^{ia}=\operatorname{softmax}_a\left(f_m\left(I_i \odot A_{i}\right)\right)$ predicted probability for class \( a \) of the \( i^{th} \) sample for the prompted model, and $p_{o}^{ia}=\operatorname{softmax}_a\left(f_o\left(I_i\right)\right)$ is for the non-prompted model. In conclusion, considering the constraints imposed by the weight-learning function, the learning objective of our study can be formally expressed as
\begin{equation}
\begin{aligned}
& \text{minimize} \quad \mathcal{L}_{\text{Pred}} + \lambda_1\mathcal{L_{\text{Param}}}  + \lambda_2\mathcal{L_{\text{Activ}}}
\\
& \text{subject to} \quad g(1) = 1
\end{aligned}
\end{equation}
where $\lambda_1$ and $\lambda_2$ are the weighting hyper-parameters for parameter sharing loss and activation sharing loss, respectively. If we treat the constraint as a Lagrange multiplier~\citep{gordon2012karush} and solve an equivalent problem by substituting the constraint to a regularization term $\mathcal{L_{\text{Agg}}}(\mathbf{\theta_{g}})$, our overall objective function can be rewritten as
\begin{equation}
\begin{aligned}
\text{minimize} \quad \mathcal{L}_{\text{Pred}} + \lambda_1 \mathcal{L_{\text{Param}}} + \lambda_2 \mathcal{L_{\text{Activ}}} + \lambda_3 \mathcal{L_{\text{Agg}}}
\end{aligned}
\end{equation}
where \(\mathcal{L}_{\text{Agg}}(\theta_{g}) = \| g(1) - 1 \|\) is commonly chosen to be the $\ell_2$-norm, $\lambda_3$ is the weighting hyper-parameters for weight-learning function constraint regularization term.

\subsubsection{Alternating Training Algorithm}
As illustrated in Figure~\ref{fig:framework}, the prompted model \( f_m \) exhibits an output-as-input structure during the prompt refinement stage. Specifically, \( f_m \)'s output is fed back into its input during the prompted prediction stage, creating circular dependencies in training~\citep{han2017alternating, xu2023convergence}. These circular dependencies may introduce challenges in the convergence of the model, potentially leading to instability or oscillatory behavior during training. To mitigate this, we propose an alternating training strategy to break this cyclic dependency and ensure better convergence. This strategy involves three models: the non-prompted model \( f_o \), the prompted model \( f_m \), and the weight-learning function \( g \).

\begin{algorithm}
\caption{Alternating Training}\label{alg:alg1}
\begin{algorithmic}[1]
\REQUIRE $I, D, y$
\ENSURE $f_m, f_o, g$
\FOR{$t = 1:T$}               
    \FOR{$q = 1:F$}
        \STATE {Compute $\nabla_{\theta_{f_m}}$ based on Equation~\ref{eq:d_grad1}}
        \STATE {Compute $\nabla_{\theta_{f_o}}$ based on Equation~\ref{eq:d_grad2}}
        \STATE {$\theta_{f_m} \gets \theta_{f_m}-\eta\nabla_{\theta_{f_m}}$}
        \STATE {$\theta_{f_o} \gets \theta_{{f_o}}-\eta \nabla_{\theta_{{f_o}}}$}
    \ENDFOR 
    \FOR{$q = 1:G$}
        \STATE {Compute $\nabla_{\theta_{g}}$ based on Equation~\ref{eq:d_grad3}}
        \STATE  {$\theta_{g} \gets \theta_{g}-\eta\nabla_{\theta_{g}}$}
    \ENDFOR
\ENDFOR
\end{algorithmic}
\end{algorithm}

The training algorithm is summarized in Algorithm~\ref{alg:alg1}. our algorithm is to fix the parameter for $g$ while updating $f_m$ and $f_o$ with a learning rate $\eta$ for $F$ iterations from Lines 2-7. The gradients w.r.t. $\theta_{f_m}$ and $\theta_{f_m}$ are computed as 
\begin{equation}
\begin{aligned}
\nabla_{f_m}=\dfrac{\partial }{\partial \theta_{f_m}}\left(\mathcal{L}_{\text{Pred}} + \lambda_1 \mathcal{L_{\text{Param}}} + \lambda_2 \mathcal{L_{\text{Activ}}}\right)
 \label{eq:d_grad1}
 \end{aligned}
\end{equation}
\begin{equation}
\begin{aligned}
\nabla_{f_o}=\dfrac{\partial }{\partial \theta_{f_o}}\left(\mathcal{L}_{\text{Pred}} + \lambda_1 \mathcal{L_{\text{Param}}} + \lambda_2 \mathcal{L_{\text{Activ}}}\right)
\label{eq:d_grad2}
\end{aligned}
\end{equation}
From Lines 8-11, our algorithm fixes the parameter for  $f_m$ and $f_o$ while updating $g$ with a learning rate $\eta$ for $G$ iterations. The gradients w.r.t. $\theta_{g}$ are computed as 
\begin{equation}
\begin{aligned}
\nabla_{\theta_{g}}=\dfrac{\partial }{\partial \theta_{g}}\left(\mathcal{L}_{\text{Pred}} + \lambda_3 \mathcal{L_{\text{Agg}}}\right)
 \label{eq:d_grad3}
 \end{aligned}
\end{equation}
We repeat the alternating training process outlined in Lines 2-11 for \( T \) iterations, continuing until optimal performance is achieved (e.g., no further increase in prediction accuracy on the validation set).

\section{Experimental Evaluation}

\begin{table*}[h]
\caption{Comparison of prediction performance on the Pancreas and LIDC datasets between our proposed framework and comparative attention-guided learning methods. The best results for each task are highlighted in boldface, and the second-best results are underlined.}
\centering
  \label{tab:results_1}
   \resizebox{0.8\linewidth}{!}{
  \begin{tabular}{c|cccc|ccccc}
    \toprule
    \multirow{2}{*}{Model} & \multicolumn{4}{c|}{Pancreas} &  \multicolumn{4}{c}{LIDC} \\
    \cline{2-9}
     & Accuracy $\uparrow$ & Precision $\uparrow$ & Recall $\uparrow$ & F1 $\uparrow$ & Accuracy $\uparrow$ & Precision $\uparrow$ & Recall $\uparrow$ & F1 $\uparrow$ \\
    \hline
    \hline    
     Baseline & 85.09 & 98.82 & 83.69 & 90.22 & 66.40 & 59.29 & 69.020 & 63.47 \\
     GRADIA  &  83.13 &  \underline{99.04}  &  81.12  & 89.10 &  67.44 & 65.65 & 73.19 & 68.99 \\
     HAICS  &  86.44  &  98.99  &  85.10  &  91.24 &  66.86 & 64.71 & 74.60 & 69.14 \\
     RES-G  &  \underline{89.89}  &  98.94  &  89.17  &  \underline{93.79} &  \underline{68.56} & \textbf{67.93} & 71.46 & 69.27 \\
     RES-L &  89.79 & 98.07 &  \underline{89.88} & 93.78 & 68.35 & 65.97 & \textbf{76.28} & \underline{70.55} \\
     VAPL & \textbf{92.31} & \textbf{99.84} & \textbf{91.03} & \textbf{95.30} & \textbf{69.45} & \underline{67.43} & \underline{75.25} & \textbf{71.13} \\
    \bottomrule
    \end{tabular}
     }
\end{table*}

\begin{table*}[h]
\caption{Comparison of prediction performance on the Gender and Scene datasets between our proposed framework and comparative attention-guided learning methods. The best results for each task are highlighted in boldface, and the second-best results are underlined.}
\centering
  \label{tab:results_2}
   \resizebox{0.8\linewidth}{!}{
  \begin{tabular}{c|cccc|ccccc}
    \toprule
    \multirow{2}{*}{Model} & \multicolumn{4}{c|}{Gender} &  \multicolumn{4}{c}{Scene} \\
    \cline{2-9}
     & Accuracy $\uparrow$ & Precision $\uparrow$ & Recall $\uparrow$ & F1 $\uparrow$ & Accuracy $\uparrow$ & Precision $\uparrow$ & Recall $\uparrow$ & F1 $\uparrow$\\
    \hline
    \hline    
   Baseline & 68.35 & 67.45 & 69.98 & 68.69 & 93.42 & 94.87 & 91.68 & 93.25 \\
   GRADIA  &  70.01 & 67.83 & 74.35 & 70.94 & 95.03 & 96.21 & 92.55 & 94.34 \\
   HAICS  &  69.29 & 66.42 & 73.61 & 69.83 & 94.89 & 95.73 & 92.94 & 94.31 \\
   RES-G  & \underline{71.33} & \underline{69.98} & \textbf{78.53} & \underline{74.01} &  \underline{95.91} & 96.22 &  \textbf{95.35} & \underline{95.78} \\
   RES-L &  70.39 & 68.41 & 73.29 & 70.77 & 95.53 &  \underline{96.98} &  \underline{94.56} & 95.75 \\
   VAPL & \textbf{73.36} & \textbf{71.43} & \underline{76.88} & \textbf{74.05} & \textbf{96.39} & \textbf{97.43} & 94.48 & \textbf{95.93}  \\
    \bottomrule
    \end{tabular}
    }
\end{table*}

\subsection{Datasets} 
To assess our framework's effectiveness, we employed four datasets: two from real-world scenarios, sourced from MS COCO~\citep{lin2014microsoft}, and two from the medical field, namely LIDC-IDRI (LIDC)~\citep{armato2011lung} and the Pancreas dataset~\citep{roth2015deeporgan}. 

Specifically, \textbf{LIDC} dataset, featuring lung CT scans with annotated lesions, was preprocessed into 2D images (224$\times$224) and augmented with noise to simulate incomplete prompts. Negative samples were created by slicing surrounding areas of nodules. The final dataset included 2625 nodules and 65505 non-nodules images, split into 100/1200/1200 for training, validation, and testing to reflect limited access to human explanations. \textbf{Pancreas} dataset, with normal images from Cancer Imaging Archive and abnormal images from MSD, was preprocessed similarly to LIDC. It included 281 CT scans with tumors and 80 without. Tumor lesions were treated as inaccurate explanations, while pancreas segmentations were ground truth. Data was split into 30/30/rest for training, validation, and testing, maintaining class balance. For \textbf{Gender} classification dataset~\citep{gao2022res}, involved extracting 1,600 images from MS COCO based on captions mentioning "men" or "women". Images with both genders, multiple people, or unclear figures were excluded. For \textbf{Scene}~\citep{gao2022res} recognition, balanced nature and urban categories from Places365, again selecting 100 images for training. each image from two datasets comes with explanations that were obtained through a custom UI. Human annotations were collected for all images, and only 100 were randomly chosen for training to simulate limited access to explanations. For each sample image, the factual region of the human-explanation annotation is treated as the ``indispensable'' area, the counterfactual region is considered the ``precluded'' area, and the remaining regions are categorized as ``undecided'' areas, thereby facilitating the creation of an incomplete visual attention prompt.

\subsection{Experimental Setup}

\noindent\textbf{Comparison Methods} To evaluate the effectiveness of our proposed Visual Attention-Prompted Prediction and Learning framework (hereafter referred to as ``VAPL''), comparative studies were conducted with four prominent attention-guided learning methods, namely, GRAIDA~\citep{gao2022aligning}, HAICS~\citep{shen2021human}, RES-G, and RES-L~\citep{gao2022res}. These comparison methods were trained in accordance with the respective implementation guidelines presented in their papers. Additionally, comparisons were made with a baseline model, comprising a ResNet-18 architecture~\citep{he2016deep}, trained exclusively using prediction loss and employing original images as input.

\noindent\textbf{Implementation Details} In this study, the ResNet18 Convolutional Neural Network (CNN) architecture was uniformly employed as the backbone model across all evaluated methods. The experimental setup was standardized with a batch size of 16, and the number of perturbed masks was set to 5000. Furthermore, a pixel conversion probability of 0.1 was established. Training was conducted over 10 epochs, each comprising 5 iterations for the alternating updating phase, effectively resulting in 50 training epochs for each model. The Adam optimization algorithm \citep{kingma2014adam} was utilized with a learning rate of 0.0001. Regarding computational resources, all experiments were executed using an NVIDIA GTX 3090 GPU. To assess the classification performance of the proposed framework, conventional metrics accuracy, precision, recall, and the F1 score were employed.

\begin{table*}[h]
\caption{Ablation study on LIDC-IDRI (left) and Pancreas dataset (right).}
\centering
    \label{ablation}
    \Large
    \resizebox{0.8\linewidth}{!}{
    \begin{tabular}{c|cccc|ccccc}
    \toprule
    \multirow{2}{*}{Ablation} & \multicolumn{4}{c|}{LIDC-IDRI} &  \multicolumn{4}{c}{Pancreas} \\
    \cline{2-9}
     & Accuracy $\uparrow$ & Precision $\uparrow$ & Recall $\uparrow$ & F1 $\uparrow$ & Accuracy $\uparrow$ & Precision $\uparrow$ & Recall $\uparrow$ & F1 $\uparrow$\\
    \hline
     \textbf{VAPL} & 69.45 & 67.43 & 75.25 & 71.13 & 92.31  & 99.84 & 91.03 & 95.30 \\
     \hline
     \textbf{VAPL-1} & 67.18 & 64.66 & 75.79 & 69.78 & 89.25 & 98.14 & 85.76 & 91.53 \\
     \textbf{VAPL-2} & 66.68 & 64.34 & 72.41 & 68.91 & 88.49 & 97.17 & 82.03 & 88.96 \\
     \textbf{VAPL-3} & 66.43 & 63.15 & 72.23 & 67.38 & 88.26 & 96.14 & 81.43 & 88.17 \\
     \textbf{VAPL-4} & 68.14 & 64.75 & 73.43 & 68.81 & 90.46 & 98.54 & 89.59 & 93.85 \\
    \bottomrule
  \end{tabular}
   }
\end{table*}

\subsection{Results Evaluation}
Table~\ref{tab:results_1} shows the quantitative evaluation of two medical image classification tasks: LIDC and Pancreas. Overall, our method outperforms all other explanation-guided learning methods on both LIDC and pancreas datasets. In particular, our model achieves the best accuracy and F1 on the pulmonary nodule classification task and the best accuracy, precision, recall, and F1 on the pancreatic tumor classification task. Our method achieved an accuracy of 92.30\% and 69.45\% on the two respective datasets, representing an improvement of 8.5\% and 4.6\% compared to the baseline Resnet-18 approach. In addition, our method attained F-1 scores of 95.30\% and 71.13\% on the two datasets, marking an enhancement of 5.6\% and 12.1\% respectively when compared to the baseline Resnet-18 method. Our findings demonstrate that the performance of our prediction methodology is enhanced when incorporating explanations into the prediction phase. While the precision or recall scores reported on the LIDC dataset are marginally lower, our approach achieves the highest F1 scores across both datasets. This outcome signifies that our overall performance surpasses that of all other attention-guided learning methods. 

Table~\ref{tab:results_2} presents a comprehensive comparison of various models across two image classification tasks: Gender and Scene. Notably, the VAPL model demonstrates superior performance, achieving the highest scores in almost all metrics for both tasks, with particularly outstanding results in the Scene task where it achieves an impressive 96.39\% in Accuracy and 97.43\% in Precision. The RES-G model also shows commendable performance, especially in the Scene task, where it records the highest Recall of 95.35\% and a competitive F1 score of 95.78\%. In contrast, the Baseline model, as expected, trails behind in most metrics, underscoring the effectiveness of advanced models like VAPL and RES-G. This analysis underscores the significance of specialized models in enhancing classification accuracy in attention-guided learning tasks, with VAPL excelling in precision-oriented metrics and RES-G showing strength in recall-focused areas.

\begin{figure}[h]
\centering
\includegraphics[width=\linewidth]{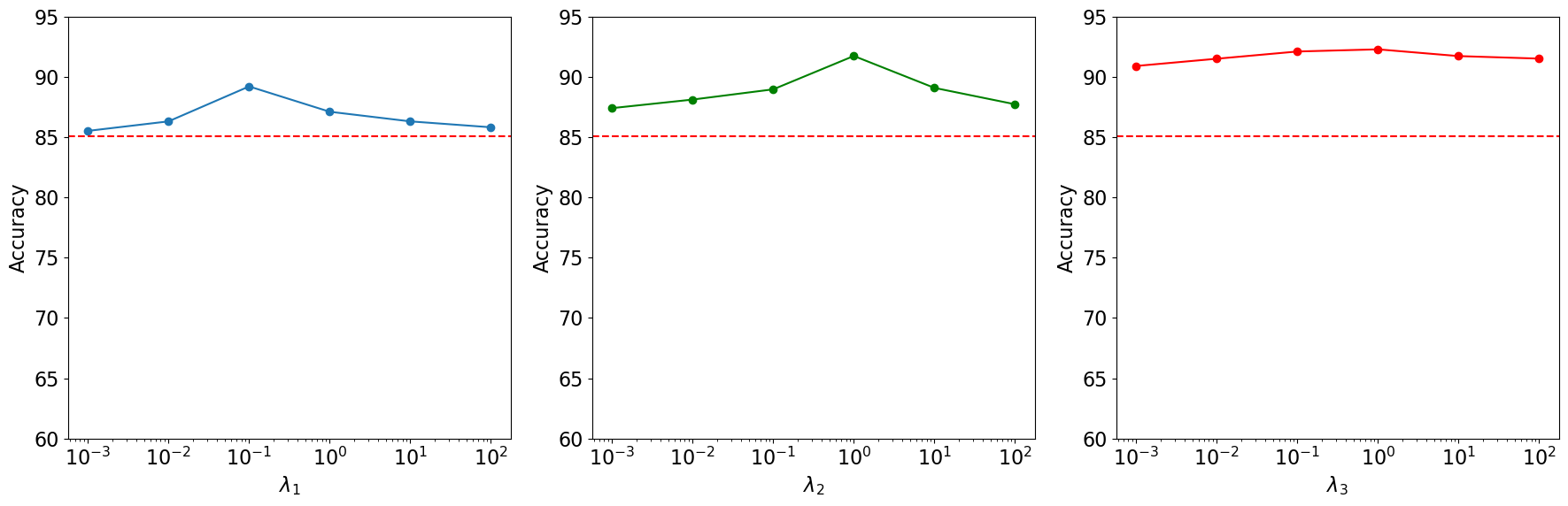}
\caption{Sensitivity Analysis on the Pancreas dataset.}
\label{fig:sensitivity}
\end{figure}

We further provide a sensitivity analysis of the hyper-parameters \(\lambda_1\), \(\lambda_2\), and \(\lambda_3\), which denote the weights of the parameter sharing regularization loss, activation sharing regularization loss, and weight-learning function regularization loss, respectively. Figure~\ref{fig:sensitivity} illustrates the accuracy of the proposed model for various weights in the context of pancreatic tumor classification, with the red dashed lines representing the baseline model’s performance. Overall, the model exhibits relatively high sensitivity to variations in the weights for the parameter sharing and activation sharing regularization losses, while demonstrating lower sensitivity to the weight-learning function regularization loss. Additionally, a concave curvature is observed for all three hyper-parameters. Optimal overall performance is achieved when \(\lambda_1\), \(\lambda_2\), and \(\lambda_3\) are set to 0.1, 1, and 10, respectively.

\subsection{Ablation Study}

To evaluate the efficacy of the proposed visual attention-prompted learning and prediction (VAPL) framework, an ablation study was conducted with four distinct variants: (\textbf{VAPL-1}), which involves the omission of the visual attention prompt refinement architecture, thereby relying exclusively on the incomplete visual attention prompt during the learning and prediction phases. (\textbf{VAPL-2}), which pertains to the exclusion of parameter-sharing regularization between the two models, while retaining co-activation regularization during training. (\textbf{VAPL-3}), which entails the removal of co-activation regularization, and maintaining only parameter-sharing regularization during the training process. (\textbf{VAPL-4}), characterized by the absence of the proposed weight-learning function and instead employing a weighted averaging method based solely on confidence scores computed by classifiers.

The outcomes of the ablation study conducted on the LIDC and Pancreas datasets are presented in Table \ref{ablation}. While each component individually contributes to the overall performance of the model, it is observed from \textbf{VAPL-2} and \textbf{VAPL-3} that co-training with parameter and activation sharing regularization offers a notably larger enhancement in predictability. Furthermore, the results about \textbf{VAPL-1} highlight the significant contribution of the proposed visual attention prompt refinement method to improved performance. This is particularly evident in the Pancreas dataset, which demonstrates a greater prevalence of prompt incompleteness issues.

\section{Conclusion and Discussion}
This research paper introduces the Visual Attention Prompted Prediction and Learning Framework, a novel approach that integrates visual attention prompts into the decision-making process of models. The framework effectively tackles challenges like incomplete information from visual prompts and predictions for samples lacking such prompts through an attention-prompted co-training mechanism with parameter-sharing and co-activation regularization. This helps align activation patterns and facilitates knowledge transfer between prompted and non-prompted models. Additionally, the framework incorporates a prompt-guided refinement method with an adaptively learnable mask aggregation function to manage prompt incompleteness. Its efficacy is validated across four datasets from various domains, showing improved predictive accuracy for samples with and without attention prompts. Despite the aforementioned advantages, we also acknowledge that prompts from users can introduce bias, though our framework is tolerant of it. In the future, we plan to explore the application of our proposed method to graph learning and prediction, incorporating prompts with graph attention mechanisms, such as subgraphs.

\section*{Acknowledgments}
This work is supported by the National Science Foundation (NSF) Grant No. 1755850, No. 1841520, No. 2007716, No. 2007976, No. 1942594, No. 1907805, a Jeffress Memorial Trust Award, Amazon Research Award, Oracle for Research Grant Award, Cisco Faculty Research Award, NVIDIA GPU Grant, Design Knowledge Company (subcontract number: 10827.002.120.04), CIFellowship (2021CIF-Emory-05), and the Department of Homeland Security under Grant No. 17STCIN00001.

\bibliographystyle{named}
\bibliography{ijcai24}

\begin{thebibliography}{}

\bibitem[\protect\citeauthoryear{Adadi and Berrada}{2018}]{adadi2018peeking}
Amina Adadi and Mohammed Berrada.
\newblock Peeking inside the black-box: a survey on explainable artificial intelligence (xai).
\newblock {\em IEEE access}, 6:52138--52160, 2018.

\bibitem[\protect\citeauthoryear{Armato~III \bgroup \em et al.\egroup }{2011}]{armato2011lung}
Samuel~G Armato~III, Geoffrey McLennan, Luc Bidaut, Michael~F McNitt-Gray, Charles~R Meyer, Anthony~P Reeves, Binsheng Zhao, Denise~R Aberle, Claudia~I Henschke, Eric~A Hoffman, et~al.
\newblock The lung image database consortium (lidc) and image database resource initiative (idri): a completed reference database of lung nodules on ct scans.
\newblock {\em Medical physics}, 38(2):915--931, 2011.

\bibitem[\protect\citeauthoryear{Bai \bgroup \em et al.\egroup }{2023}]{bai2023saliency}
Guangji Bai, Qilong Zhao, Xiaoyang Jiang, Yifei Zhang, and Liang Zhao.
\newblock Saliency-guided hidden associative replay for continual learning.
\newblock {\em arXiv preprint arXiv:2310.04334}, 2023.

\bibitem[\protect\citeauthoryear{Camburu \bgroup \em et al.\egroup }{2018}]{camburu2018snli}
Oana-Maria Camburu, Tim Rockt{\"a}schel, Thomas Lukasiewicz, and Phil Blunsom.
\newblock e-snli: Natural language inference with natural language explanations.
\newblock {\em Advances in Neural Information Processing Systems}, 31, 2018.

\bibitem[\protect\citeauthoryear{Chen \bgroup \em et al.\egroup }{2020}]{chen2020air}
Shi Chen, Ming Jiang, Jinhui Yang, and Qi~Zhao.
\newblock Air: Attention with reasoning capability.
\newblock In {\em Computer Vision--ECCV 2020: 16th European Conference, Glasgow, UK, August 23--28, 2020, Proceedings, Part I 16}, pages 91--107. Springer, 2020.

\bibitem[\protect\citeauthoryear{Devlin \bgroup \em et al.\egroup }{2018}]{devlin2018bert}
Jacob Devlin, Ming-Wei Chang, Kenton Lee, and Kristina Toutanova.
\newblock Bert: Pre-training of deep bidirectional transformers for language understanding.
\newblock {\em arXiv preprint arXiv:1810.04805}, 2018.

\bibitem[\protect\citeauthoryear{Dosovitskiy \bgroup \em et al.\egroup }{2020}]{dosovitskiy2020image}
Alexey Dosovitskiy, Lucas Beyer, Alexander Kolesnikov, Dirk Weissenborn, Xiaohua Zhai, Thomas Unterthiner, Mostafa Dehghani, Matthias Minderer, Georg Heigold, Sylvain Gelly, et~al.
\newblock An image is worth 16x16 words: Transformers for image recognition at scale.
\newblock {\em arXiv preprint arXiv:2010.11929}, 2020.

\bibitem[\protect\citeauthoryear{Gao \bgroup \em et al.\egroup }{2022a}]{gao2022aligning}
Yuyang Gao, Tong Sun, Liang Zhao, and Sungsoo Hong.
\newblock Aligning eyes between humans and deep neural network through interactive attention alignment, 2022.

\bibitem[\protect\citeauthoryear{Gao \bgroup \em et al.\egroup }{2022b}]{gao2022res}
Yuyang Gao, Tong~Steven Sun, Guangji Bai, Siyi Gu, Sungsoo~Ray Hong, and Zhao Liang.
\newblock Res: A robust framework for guiding visual explanation.
\newblock In {\em Proceedings of the 28th ACM SIGKDD Conference on Knowledge Discovery and Data Mining}, pages 432--442, 2022.

\bibitem[\protect\citeauthoryear{Garcia \bgroup \em et al.\egroup }{2018}]{garcia2018modality}
Nuno~C Garcia, Pietro Morerio, and Vittorio Murino.
\newblock Modality distillation with multiple stream networks for action recognition.
\newblock In {\em Proceedings of the European Conference on Computer Vision (ECCV)}, pages 103--118, 2018.

\bibitem[\protect\citeauthoryear{Gordon and Tibshirani}{2012}]{gordon2012karush}
Geoff Gordon and Ryan Tibshirani.
\newblock Karush-kuhn-tucker conditions.
\newblock {\em Optimization}, 10(725/36):725, 2012.

\bibitem[\protect\citeauthoryear{Gu \bgroup \em et al.\egroup }{2023}]{gu2023essa}
Siyi Gu, Yifei Zhang, Yuyang Gao, Xiaofeng Yang, and Liang Zhao.
\newblock Essa: Explanation iterative supervision via saliency-guided data augmentation.
\newblock In {\em Proceedings of the 29th ACM SIGKDD Conference on Knowledge Discovery and Data Mining}, pages 567--576, 2023.

\bibitem[\protect\citeauthoryear{Hajialigol \bgroup \em et al.\egroup }{2023}]{hajialigol2023xai}
Daniel Hajialigol, Hanwen Liu, and Xuan Wang.
\newblock Xai-class: Explanation-enhanced text classification with extremely weak supervision.
\newblock {\em arXiv preprint arXiv:2311.00189}, 2023.

\bibitem[\protect\citeauthoryear{Han \bgroup \em et al.\egroup }{2017}]{han2017alternating}
Tian Han, Yang Lu, Song-Chun Zhu, and Ying~Nian Wu.
\newblock Alternating back-propagation for generator network.
\newblock In {\em Proceedings of the AAAI Conference on Artificial Intelligence}, volume~31, 2017.

\bibitem[\protect\citeauthoryear{He \bgroup \em et al.\egroup }{2016}]{he2016deep}
Kaiming He, Xiangyu Zhang, Shaoqing Ren, and Jian Sun.
\newblock Deep residual learning for image recognition.
\newblock In {\em Proceedings of the IEEE conference on computer vision and pattern recognition}, pages 770--778, 2016.

\bibitem[\protect\citeauthoryear{Hsieh \bgroup \em et al.\egroup }{2023}]{hsieh2023distilling}
Cheng-Yu Hsieh, Chun-Liang Li, Chih-Kuan Yeh, Hootan Nakhost, Yasuhisa Fujii, Alexander Ratner, Ranjay Krishna, Chen-Yu Lee, and Tomas Pfister.
\newblock Distilling step-by-step! outperforming larger language models with less training data and smaller model sizes.
\newblock {\em arXiv preprint arXiv:2305.02301}, 2023.

\bibitem[\protect\citeauthoryear{Jia \bgroup \em et al.\egroup }{2022}]{jia2022visual}
Menglin Jia, Luming Tang, Bor-Chun Chen, Claire Cardie, Serge Belongie, Bharath Hariharan, and Ser-Nam Lim.
\newblock Visual prompt tuning.
\newblock In {\em European Conference on Computer Vision}, pages 709--727. Springer, 2022.

\bibitem[\protect\citeauthoryear{Kingma and Ba}{2014}]{kingma2014adam}
Diederik~P Kingma and Jimmy Ba.
\newblock Adam: A method for stochastic optimization.
\newblock {\em arXiv preprint arXiv:1412.6980}, 2014.

\bibitem[\protect\citeauthoryear{Li \bgroup \em et al.\egroup }{2023}]{li2023boosting}
Hao Li, Dingwen Zhang, Nian Liu, Lechao Cheng, Yalun Dai, Chao Zhang, Xinggang Wang, and Junwei Han.
\newblock Boosting low-data instance segmentation by unsupervised pre-training with saliency prompt.
\newblock In {\em Proceedings of the IEEE/CVF Conference on Computer Vision and Pattern Recognition}, pages 15485--15494, 2023.

\bibitem[\protect\citeauthoryear{Lin \bgroup \em et al.\egroup }{2014}]{lin2014microsoft}
Tsung-Yi Lin, Michael Maire, Serge Belongie, James Hays, Pietro Perona, Deva Ramanan, Piotr Doll{\'a}r, and C~Lawrence Zitnick.
\newblock Microsoft coco: Common objects in context.
\newblock In {\em European conference on computer vision}, pages 740--755. Springer, 2014.

\bibitem[\protect\citeauthoryear{Liu \bgroup \em et al.\egroup }{2023}]{liu2023kept}
Jintao Liu, Zequn Zhang, Zhi Guo, Li~Jin, Xiaoyu Li, Kaiwen Wei, and Xian Sun.
\newblock Kept: Knowledge enhanced prompt tuning for event causality identification.
\newblock {\em Knowledge-Based Systems}, 259:110064, 2023.

\bibitem[\protect\citeauthoryear{Lopez-Paz \bgroup \em et al.\egroup }{2015}]{lopez2015unifying}
David Lopez-Paz, L{\'e}on Bottou, Bernhard Sch{\"o}lkopf, and Vladimir Vapnik.
\newblock Unifying distillation and privileged information.
\newblock {\em arXiv preprint arXiv:1511.03643}, 2015.

\bibitem[\protect\citeauthoryear{Ma \bgroup \em et al.\egroup }{1994}]{279274}
Changxue Ma, Y.~Kamp, and L.F. Willems.
\newblock A frobenius norm approach to glottal closure detection from the speech signal.
\newblock {\em IEEE Transactions on Speech and Audio Processing}, 2(2):258--265, 1994.

\bibitem[\protect\citeauthoryear{Montavon \bgroup \em et al.\egroup }{2019}]{montavon2019layer}
Gr{\'e}goire Montavon, Alexander Binder, Sebastian Lapuschkin, Wojciech Samek, and Klaus-Robert M{\"u}ller.
\newblock Layer-wise relevance propagation: an overview.
\newblock {\em Explainable AI: interpreting, explaining and visualizing deep learning}, pages 193--209, 2019.

\bibitem[\protect\citeauthoryear{Narang \bgroup \em et al.\egroup }{2020}]{narang2020wt5}
Sharan Narang, Colin Raffel, Katherine Lee, Adam Roberts, Noah Fiedel, and Karishma Malkan.
\newblock Wt5?! training text-to-text models to explain their predictions.
\newblock {\em arXiv preprint arXiv:2004.14546}, 2020.

\bibitem[\protect\citeauthoryear{Nguyen \bgroup \em et al.\egroup }{2023}]{nguyen2023mononet}
An-Phi Nguyen, Dana~Lea Moreno, Nicolas Le-Bel, and Mar{\'\i}a Rodr{\'\i}guez~Mart{\'\i}nez.
\newblock Mononet: Enhancing interpretability in neural networks via monotonic features.
\newblock {\em Bioinformatics Advances}, 3(1):vbad016, 2023.

\bibitem[\protect\citeauthoryear{Oymak \bgroup \em et al.\egroup }{2023}]{oymak2023role}
Samet Oymak, Ankit~Singh Rawat, Mahdi Soltanolkotabi, and Christos Thrampoulidis.
\newblock On the role of attention in prompt-tuning.
\newblock {\em arXiv preprint arXiv:2306.03435}, 2023.

\bibitem[\protect\citeauthoryear{Paiss \bgroup \em et al.\egroup }{2022}]{paiss2022no}
Roni Paiss, Hila Chefer, and Lior Wolf.
\newblock No token left behind: Explainability-aided image classification and generation.
\newblock In {\em European Conference on Computer Vision}, pages 334--350. Springer, 2022.

\bibitem[\protect\citeauthoryear{Pan \bgroup \em et al.\egroup }{2024}]{pan2024distilling}
Bo~Pan, Zheng Zhang, Yifei Zhang, Yuntong Hu, and Liang Zhao.
\newblock Distilling large language models for text-attributed graph learning.
\newblock {\em arXiv preprint arXiv:2402.12022}, 2024.

\bibitem[\protect\citeauthoryear{Qi \bgroup \em et al.\egroup }{2019}]{qi2019visualizing}
Zhongang Qi, Saeed Khorram, and Fuxin Li.
\newblock Visualizing deep networks by optimizing with integrated gradients.
\newblock In {\em CVPR Workshops}, volume~2, pages 1--4, 2019.

\bibitem[\protect\citeauthoryear{Raffel \bgroup \em et al.\egroup }{2020}]{raffel2020exploring}
Colin Raffel, Noam Shazeer, Adam Roberts, Katherine Lee, Sharan Narang, Michael Matena, Yanqi Zhou, Wei Li, and Peter~J Liu.
\newblock Exploring the limits of transfer learning with a unified text-to-text transformer.
\newblock {\em The Journal of Machine Learning Research}, 21(1):5485--5551, 2020.

\bibitem[\protect\citeauthoryear{Roth \bgroup \em et al.\egroup }{2015}]{roth2015deeporgan}
Holger~R Roth, Le~Lu, Amal Farag, Hoo-Chang Shin, Jiamin Liu, Evrim~B Turkbey, and Ronald~M Summers.
\newblock Deeporgan: Multi-level deep convolutional networks for automated pancreas segmentation.
\newblock In {\em Medical Image Computing and Computer-Assisted Intervention--MICCAI 2015: 18th International Conference, Munich, Germany, October 5-9, 2015, Proceedings, Part I 18}, pages 556--564. Springer, 2015.

\bibitem[\protect\citeauthoryear{Selvaraju \bgroup \em et al.\egroup }{2017}]{selvaraju2017grad}
Ramprasaath~R Selvaraju, Michael Cogswell, Abhishek Das, Ramakrishna Vedantam, Devi Parikh, and Dhruv Batra.
\newblock Grad-cam: Visual explanations from deep networks via gradient-based localization.
\newblock In {\em Proceedings of the IEEE international conference on computer vision}, pages 618--626, 2017.

\bibitem[\protect\citeauthoryear{Shen \bgroup \em et al.\egroup }{2021}]{shen2021human}
Haifeng Shen, Kewen Liao, Zhibin Liao, Job Doornberg, Maoying Qiao, Anton Van Den~Hengel, and Johan~W Verjans.
\newblock Human-ai interactive and continuous sensemaking: A case study of image classification using scribble attention maps.
\newblock In {\em Extended Abstracts of CHI}, pages 1--8, 2021.

\bibitem[\protect\citeauthoryear{Vapnik \bgroup \em et al.\egroup }{2015}]{vapnik2015learning}
Vladimir Vapnik, Rauf Izmailov, et~al.
\newblock Learning using privileged information: similarity control and knowledge transfer.
\newblock {\em J. Mach. Learn. Res.}, 16(1):2023--2049, 2015.

\bibitem[\protect\citeauthoryear{Vaswani \bgroup \em et al.\egroup }{2017}]{vaswani2017attention}
Ashish Vaswani, Noam Shazeer, Niki Parmar, Jakob Uszkoreit, Llion Jones, Aidan~N Gomez, {\L}ukasz Kaiser, and Illia Polosukhin.
\newblock Attention is all you need.
\newblock {\em Advances in neural information processing systems}, 30, 2017.

\bibitem[\protect\citeauthoryear{Xu \bgroup \em et al.\egroup }{2023}]{xu2023convergence}
Jintao Xu, Chenglong Bao, and Wenxun Xing.
\newblock Convergence rates of training deep neural networks via alternating minimization methods.
\newblock {\em Optimization Letters}, pages 1--15, 2023.

\bibitem[\protect\citeauthoryear{Ye \bgroup \em et al.\egroup }{2022}]{ye2022ontology}
Hongbin Ye, Ningyu Zhang, Shumin Deng, Xiang Chen, Hui Chen, Feiyu Xiong, Xi~Chen, and Huajun Chen.
\newblock Ontology-enhanced prompt-tuning for few-shot learning.
\newblock In {\em Proceedings of the ACM Web Conference 2022}, pages 778--787, 2022.

\bibitem[\protect\citeauthoryear{Zhang \bgroup \em et al.\egroup }{2023}]{zhang2023magi}
Yifei Zhang, Siyi Gu, Yuyang Gao, Bo~Pan, Xiaofeng Yang, and Liang Zhao.
\newblock Magi: Multi-annotated explanation-guided learning.
\newblock In {\em Proceedings of the IEEE/CVF International Conference on Computer Vision}, pages 1977--1987, 2023.

\bibitem[\protect\citeauthoryear{Zhou \bgroup \em et al.\egroup }{2015}]{zhou2015learning}
Bolei Zhou, Aditya Khosla, Agata Lapedriza, Aude Oliva, and Antonio Torralba.
\newblock Learning deep features for discriminative localization, 2015.

\end{thebibliography}

\end{document}